\documentclass[nohyperref]{article}

\usepackage{microtype}
\usepackage{graphicx}
\usepackage{subcaption}
\usepackage{booktabs} 
\usepackage{soul}
\usepackage{hyperref}

\usepackage[accepted]{icml2022}

\usepackage{amsmath}
\usepackage{amssymb}
\usepackage{mathtools}
\usepackage{amsthm}
\usepackage{subcaption}
\usepackage{booktabs}
\usepackage{epsfig}
\usepackage{setspace}
\usepackage{comment}
\usepackage{xcolor}
\usepackage{multirow}
\usepackage{enumitem} 
\usepackage[capitalize,noabbrev]{cleveref}

\theoremstyle{plain}
\newtheorem{theorem}{Theorem}[section]
\newtheorem{proposition}[theorem]{Proposition}

\theoremstyle{definition}

\theoremstyle{remark}

\usepackage[textsize=tiny]{todonotes}

\icmltitlerunning{Mitigating Neural Network Overconfidence with Logit Normalization}

\begin{document}

\twocolumn[
\icmltitle{Mitigating Neural Network Overconfidence with Logit Normalization}

\icmlsetsymbol{equal}{*}

\begin{icmlauthorlist}
\icmlauthor{Hongxin Wei}{ntu}
\icmlauthor{Renchunzi Xie}{ntu}
\icmlauthor{Hao Cheng}{ntu,nju}
\icmlauthor{Lei Feng}{cqu}
\icmlauthor{Bo An}{ntu}
\icmlauthor{Yixuan Li}{um}
\end{icmlauthorlist}

\icmlaffiliation{ntu}{Nanyang Technological University, Singapore}
\icmlaffiliation{nju}{Nanjing University, Nanjing, Jiangsu, China}
\icmlaffiliation{um}{University of Wisconsin-Madison, Wisconsin, United States}
\icmlaffiliation{cqu}{Chongqing University, Chongqing, China}

\icmlcorrespondingauthor{Renchunzi Xie}{XIER0002@e.ntu.edu.sg}

\icmlkeywords{Machine Learning, ICML}

\vskip 0.3in
]

\printAffiliationsAndNotice{} 

\begin{abstract}
Detecting out-of-distribution inputs is critical for the safe deployment of machine learning models in the real world. However, neural networks are known to suffer from the overconfidence issue, where they produce abnormally high confidence for both in- and out-of-distribution inputs. 
In this work, we show that this issue can be mitigated through Logit Normalization  (\textbf{LogitNorm})---a simple fix to the cross-entropy loss---by enforcing a constant vector norm on the logits in training. Our method is motivated by the analysis that the norm of the logit keeps increasing during training, leading to overconfident output. Our key idea behind LogitNorm is thus to decouple the influence of output’s norm during network optimization. Trained with LogitNorm, neural networks produce highly distinguishable confidence scores between in- and out-of-distribution data. Extensive experiments demonstrate the superiority of LogitNorm, reducing the average FPR95 by up to 42.30\% on common benchmarks.

\end{abstract}

\section{Introduction}
\label{introduction}

Modern neural networks deployed in the open world often struggle with out-of-distribution (OOD) inputs---samples from a different distribution that the network has not been exposed to during training, and therefore should not be predicted with high confidence at test time. A reliable classifier should not only accurately classify known in-distribution (ID) samples, but also identify as ``unknown'' any OOD input. 
This gives rise to the importance of OOD detection, which determines whether an input is ID or OOD and allows the
model to take precautions in deployment. 

A naive solution uses the maximum softmax probability (MSP)---also known as the softmax confidence score---for OOD detection~\cite{hendrycks2016baseline}. The operating hypothesis is that OOD data should trigger relatively lower softmax confidence than that of ID data. Whilst intuitive, the reality shows a non-trivial dilemma. 
In particular, deep neural networks can easily produce overconfident predictions, \emph{i.e.}, abnormally high softmax confidences, even when the inputs are far away from the training data~\cite{nguyen2015deep}. This has cast significant doubt on using softmax confidence for OOD detection. Indeed, many prior works turned to define alternative OOD scoring functions \cite{liang2018enhancing, lee2018simple, liu2020energy, sastry2020detecting, sun2021react, huang2021importance, sun2022knn}. Yet to date, the community still has a limited understanding of the fundamental cause and mitigation of the overconfidence issue.

In this work, we show that the overconfidence issue can be mitigated through a simple fix to the cross-entropy loss---the most commonly used training objective for classification---by enforcing a constant norm on the logit vector (\emph{i.e.}, pre-softmax output). Our method, \emph{Logit Normalization} (dubbed \textbf{LogitNorm}), is motivated by our analysis on the norm of the neural network's logit vectors. We find that even when most training examples are classified to their correct labels, the softmax cross-entropy loss can continue to increase the magnitude of the logit vectors. The growing magnitude during training thus leads to the overconfidence issue, despite having no improvement on the classification accuracy.  

To mitigate the issue, our key idea behind LogitNorm is to decouple the influence of output's norm from the training objective and its optimization. This can be achieved by normalizing the logit vector to have a constant norm during training. In effect, our LogitNorm loss encourages the direction of the logit output to be consistent with the corresponding one-hot label, without exacerbating the magnitude of the output. Trained with normalized outputs, the network tends to give conservative predictions and results in strong separability of softmax confidence scores between ID and OOD inputs (see Figure~\ref{fig:pdf}).

Extensive experiments demonstrate the superiority of LogitNorm over existing methods for OOD detection. First, our method drastically improves the OOD detection performance using the softmax confidence score. For example, using CIFAR-10 dataset as ID and SVHN as OOD data, our approach reduces the FPR95 from 50.33\% to 8.03\%---a \textbf{42.30}\% of improvement over the baseline~\cite{hendrycks2016baseline}. Averaged over a diverse collection of OOD datasets, our method reduces the FPR95 by \textbf{33.87}\% compared to using the softmax score with cross-entropy loss. Beyond MSP, we show that our method not only outperforms, but also boosts more advanced post-hoc OOD scoring functions, such as ODIN~\cite{liang2018enhancing}, energy score~\cite{liu2020energy}, and GradNorm score~\cite{huang2021importance}. In addition to the OOD detection task, our method improves the calibration performance on the ID data itself by way of post-hoc temperature scaling. 

Overall, \emph{using LogitNorm loss achieves strong performance on OOD detection and calibration tasks while maintaining the classification accuracy on ID data.} 
Our method can be easily adopted in practice. It is straightforward to implement with existing deep learning frameworks, and does not require sophisticated changes to the loss or training scheme. Code and data are publicly available at  \url{https://github.com/hongxin001/logitnorm_ood}.

We summarize our contributions as follows:
\begin{enumerate}
\setlength\itemsep{0.005em}
    \item We introduce LogitNorm -- a simple and effective alternative to the cross-entropy loss, which decouples the influence of logits' norm from the training procedure. We show that LogitNorm can effectively generalize to different network architectures and boost different post-hoc OOD detection methods.
    \item  We conduct extensive evaluations to show that LogitNorm can improve both OOD detection and confidence calibration while maintaining the classification accuracy on ID data. Compared with the cross-entropy loss, LogitNorm achieves an FPR95 reduction of $33.87$\% on common benchmarks with the softmax confidence score.
    \item We perform ablation studies that lead to improved understandings of our method. In particular, we contrast with alternative methods (\emph{e.g.}, GODIN~\cite{hsu2020generalized}, Logit Penalty) and demonstrate the advantages of LogitNorm. We hope that our insights inspire future research to further explore loss function design for OOD detection.
\end{enumerate}

\section{Background}
\label{introduction}

\subsection{Preliminaries: Out-of-distribution Detection}

\paragraph{Setup.} We consider a supervised multi-class classification problem. We denote by $\mathcal{X}$ the input space and $\mathcal{Y} = \{1, \ldots, k\}$ the label space with $k$ classes.  The training dataset $\mathcal{D}_\text{train}=\{\boldsymbol{x}_i, y_i\}^N_{i=1}$ consists of $N$ data points, sampled \emph{i.i.d.} from a joint data distribution $\mathcal{P}_{\mathcal{X}\mathcal{Y}}$. We use $\mathcal{P}_{\text{in}}$ to denote the marginal distribution over $\mathcal{X}$, which represents the in-distribution (ID). Given the training dataset, we learn a classifier $f: \mathcal{X} \rightarrow \mathbb{R}^k$ with trainable parameter ${\theta} \in \mathbb{R}^p$, which maps an input to the output space. An ideal classifier can be obtained by minimizing the following expected risk:
\begin{equation*}
\mathcal{R}_{\mathcal{L}}(f) = \mathbb{E}_{(\boldsymbol{x},y)\sim\mathcal{P}_{\mathcal{X}\mathcal{Y}}}\left[\mathcal{L}\left(f(\boldsymbol{x} ; {\theta}), y\right)\right],
\end{equation*}
where $\mathcal{L}$ is the commonly used cross-entropy loss with the softmax activation function:
\begin{align}
\nonumber
\mathcal{L}_{\text{CE}}(f(\boldsymbol{x};{\theta}),y) = -\log p(y|\boldsymbol{x})
&= - \log \frac{e^{f_y(\boldsymbol{x}; {\theta})}}{\sum^{k}_{i=1} e^{f_{i}(\boldsymbol{x}; {\theta})}}.
\end{align}
Here, $f_y(\boldsymbol{x}; {\theta})$ denotes the $y$-th element of $f(\boldsymbol{x}; {\theta})$ corresponding to the ground-truth label $y$, and $p(y|\boldsymbol{x})$ is the corresponding softmax probability.

\textbf{Problem statement.} During the deployment time, it is ideal that the test data are drawn from the same distribution $\mathcal{P}_\text{in}$ as the training data. However, in reality, inputs from unknown distributions can arise, whose label set may have no intersection with $\mathcal{Y}$. Such inputs are termed out-of-distribution (OOD) data and should not be predicted by the model. 

The OOD detection task can be formulated as a binary classification problem: determining whether an input $\boldsymbol{x} \in \mathcal{X}$ is from $\mathcal{P}_{\text{in}}$ (ID) or not (OOD). 
OOD detection can be performed by a level-set estimation:
\begin{equation}
g(\boldsymbol{x})= \begin{cases}\text { in } & \text { if } S(\boldsymbol{x}) \geq \gamma \\ \text { out} & \text { if } S(\boldsymbol{x})<\gamma\end{cases},
\end{equation}
where $S(\boldsymbol{x})$ denotes a scoring function and $\gamma$ is commonly chosen so that a high fraction (e.g., 95\%) of ID data is correctly classified. By convention, samples with higher scores $S(\boldsymbol{x})$ are classified as ID and vice versa. In Section~\ref{sec:results}, we will consider a variety of popular OOD scoring functions including MSP~\cite{hendrycks2016baseline}, ODIN~\cite{liang2018enhancing}, energy score~\cite{liu2020energy} and GradNorm \cite{huang2021importance}.

\section{Method: Logit Normalization}
\label{sec:method}

\subsection{Motivation}

In the following, we investigate why neural networks trained with the common softmax cross-entropy loss tend to give overconfident predictions. Our analysis suggests that the large magnitude of neural network output can be a culprit.

For notation shorthand, we use $\boldsymbol{f}$ to denote the network output $f(\boldsymbol{x}; {\theta})$ for an input $\boldsymbol{x}$. $f(\boldsymbol{x}; {\theta})$ is also known as the logit, or pre-softmax output. Without loss of generality, the logit vector $\boldsymbol{f}$ can be decomposed into two components:
\begin{equation}
\label{eq:decouple}
\boldsymbol{f} = \|\boldsymbol{f}\| \cdot \boldsymbol{\hat{f}},
\end{equation}
where $\|\boldsymbol{f}\| =\sqrt{\boldsymbol{f}_1^2 + \boldsymbol{f}_2^2 + \cdots + \boldsymbol{f}_k^2}$
is the Euclidean norm of the logit vector $\|\boldsymbol{f}\|$, and $\boldsymbol{\hat{f}}$ is the unit vector in the same direction as $\boldsymbol{f}$. In other word, $\|\boldsymbol{f}\|$ and $\boldsymbol{\hat{f}}$ represent the \emph{magnitude} and the \emph{direction} of the logit vector $\boldsymbol{f}$, respectively.

During the test stage, the model makes class predictions by $c = \arg \max_{i}(f_i)$. We have the following propositions.

\begin{proposition}
\label{prop:magnitude}
For any give constant value $s > 1$, if $\arg \max_{i} (f_i) = c$, then $\arg \max_{i} (sf_i) = c$ always holds.
\end{proposition}

Given the above proposition, we find that scaling the magnitude $\|\boldsymbol{f}\|$ of the logit does not change the predicted class $c$. In the following, we further explore how it impacts the softmax confidence score.

\begin{proposition}
\label{prop:scale}
For the softmax cross-entropy loss, let $\sigma$ be the softmax activation function. For any given scalar $s > 1$, if $c =
\arg \max_{i} (f_i)$, then $\sigma_c(s\boldsymbol{f}) \geq \sigma_c(\boldsymbol{f})$ holds.
\end{proposition}

The proofs of the above propositions are presented in Appendix~\ref{app:proofs_1} and \ref{app:proofs_2}. From Proposition \ref{prop:scale}, we find that increasing the magnitude $\|\boldsymbol{f}\|$ will cause a higher value for the softmax confidence score but leave the final prediction unchanged.

To analyze the impact on training objective, we provide the following formulation according to Eq.~(\ref{eq:decouple}):
\begin{gather}
\nonumber
\mathcal{L}_{\text{CE}}(f(\boldsymbol{x};{\theta}),y) = -\log p(y|\boldsymbol{x})
= - \log \frac{e^{\|\boldsymbol{f}\| \cdot {\hat{f}_y}}}{\sum^{k}_{i=1}e^{\|\boldsymbol{f}\| \cdot {\hat{f}_i}}}.
\end{gather}
We can find that the training loss depends on the magnitude $\|\boldsymbol{f}\|$ and the direction $\boldsymbol{\hat{f}}$. By keeping the direction unchanged, we analyze the influence of the magnitude $\|\boldsymbol{f}\|$ on the training loss. When $y =
\arg \max_{i} (f_i)$, we can see that increasing $\|\boldsymbol{f}\|$ would increase $p(y|\boldsymbol{x})$. It implies that, for those training examples that are already classified correctly, the optimization on the training loss would further increase the magnitude $\|\boldsymbol{f}\|$ of the network output to produce a higher softmax confidence score, thus obtaining a smaller loss. 

\begin{figure}[h]
    \centering
    \includegraphics[width=0.35\textwidth]{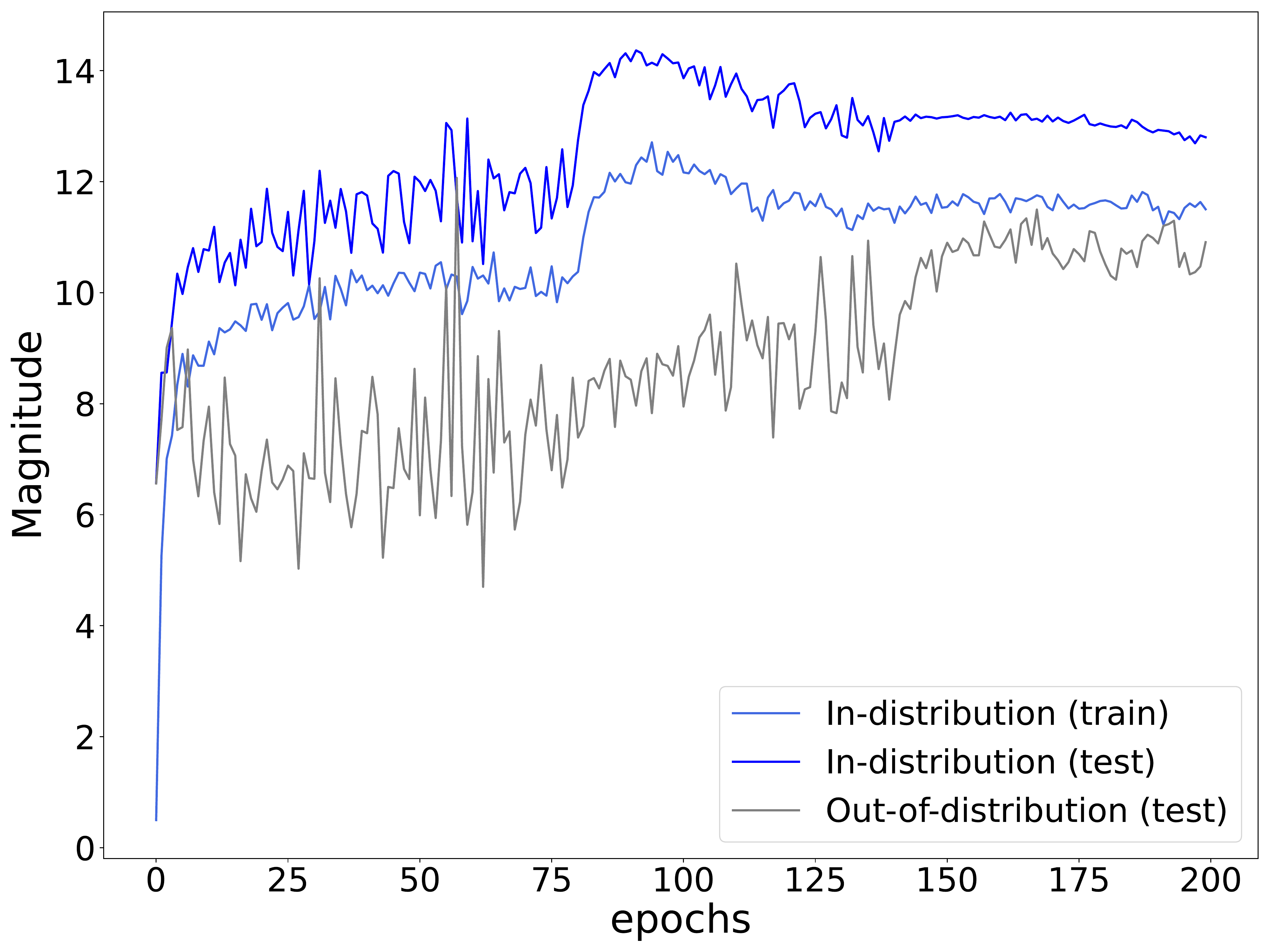}
    \vspace{-0.3cm}
    \caption{The mean magnitudes of logits under different training epochs. Model is trained on CIFAR-10 with WRN-40-2~\cite{zagoruyko2016wide}. OOD examples are from SVHN dataset.}
    \label{fig:norm}
    \vspace{-5pt}
\end{figure}

To provide a straightforward view, we show in Figure~\ref{fig:norm} the dynamics of logit norm during training. Indeed, softmax cross-entropy loss encourages the model to produce logits with increasingly larger norms for both ID and OOD examples. The large norms directly translate into overconfident softmax scores, leading to difficulty in separating ID vs. OOD data. We proceed by introducing our method, targeting this problem.

\subsection{Method}

In our previous analysis, we show that the softmax cross-entropy loss encourages the network to produce logits with larger magnitudes, leading to the overconfidence issue that makes it difficult to distinguish ID and OOD examples. To alleviate this issue, our key idea is to decouple the influence of logits' magnitude from network optimization. In other words, our goal is to keep the $L_2$ vector norm of logits a constant during the training.  
Formally, the objective can be formulated as:
\begin{align*}
&\text{minimize} \quad \mathbb{E}_{\mathcal{P}_{\mathcal{X}\mathcal{Y}}}\left[\mathcal{L}_{\text{CE}}\left(f(\boldsymbol{x} ; {\theta}), y\right)\right] \\
&\text{subject to} \quad \left\|f(\boldsymbol{x} ; {\theta})\right\|_{2}=\alpha.
\end{align*}

Performing constrained optimization in the context of modern neural networks is non-trivial. As we will show later in Section~\ref{sec:discussion}, simply adding the constraint via Lagrange multiplier \cite{forst2010optimization} does not work well. To circumvent the issue, we convert the objective into an alternative loss function that can be end-to-end trainable, which strictly enforces a constant vector norm.  

\textbf{Logit Normalization.} 
We employ \emph{Logit Normalization} (dubbed LogitNorm), which encourages the direction of the logit to be consistent with the corresponding one-hot label, without optimizing the magnitude of the logit. In particular, the logit vector is normalized to be a unit vector with a constant magnitude. The softmax cross-entropy loss is then applied on the normalized logit vector instead of the original output. Formally, the objective function of LogitNorm is given by:
\begin{equation}
    \mathcal{R}_{\mathcal{L}}(f) = \mathbb{E}_{(\boldsymbol{x},y)\sim\mathcal{P}_{\mathcal{X}\mathcal{Y}}}\left[\mathcal{L}_{\text{CE}}\left({\hat{f}}(\boldsymbol{x};\theta), y\right)\right],
\end{equation}
where ${\hat{f}}(\boldsymbol{x};\theta) = f(\boldsymbol{x};\theta)/ \|f(\boldsymbol{x};\theta)\|$ is the normalized logit vector. In practice, a small positive value (e.g., $10^{-7}$) is added to the denominator to ensure numerical stability. 
Equivalently, the new loss function can be defined as:
\begin{align}
\label{eq:norm_loss}
\mathcal{L}_{\text{logit\_norm}}({f}(\boldsymbol{x};\theta),y) = - \log \frac{e^{{f}_y/(\tau \|\boldsymbol{f}\|)}}{\sum^{k}_{i=1} e^{f_{i}/(\tau \|\boldsymbol{f}\|)}},
\end{align}
where the temperature parameter $\tau$ modulates the magnitude of the logits. Interestingly, our loss function can be viewed as having an \emph{input-dependent} temperature, $\tau \lVert f(\boldsymbol{x};\theta)\rVert$, which depends on the input $\boldsymbol{x}$.  

\begin{figure}[t]
    \centering
    \begin{subfigure}[b]{0.22\textwidth}
        \centering
        \includegraphics[height=3.0cm,width=3.8cm]{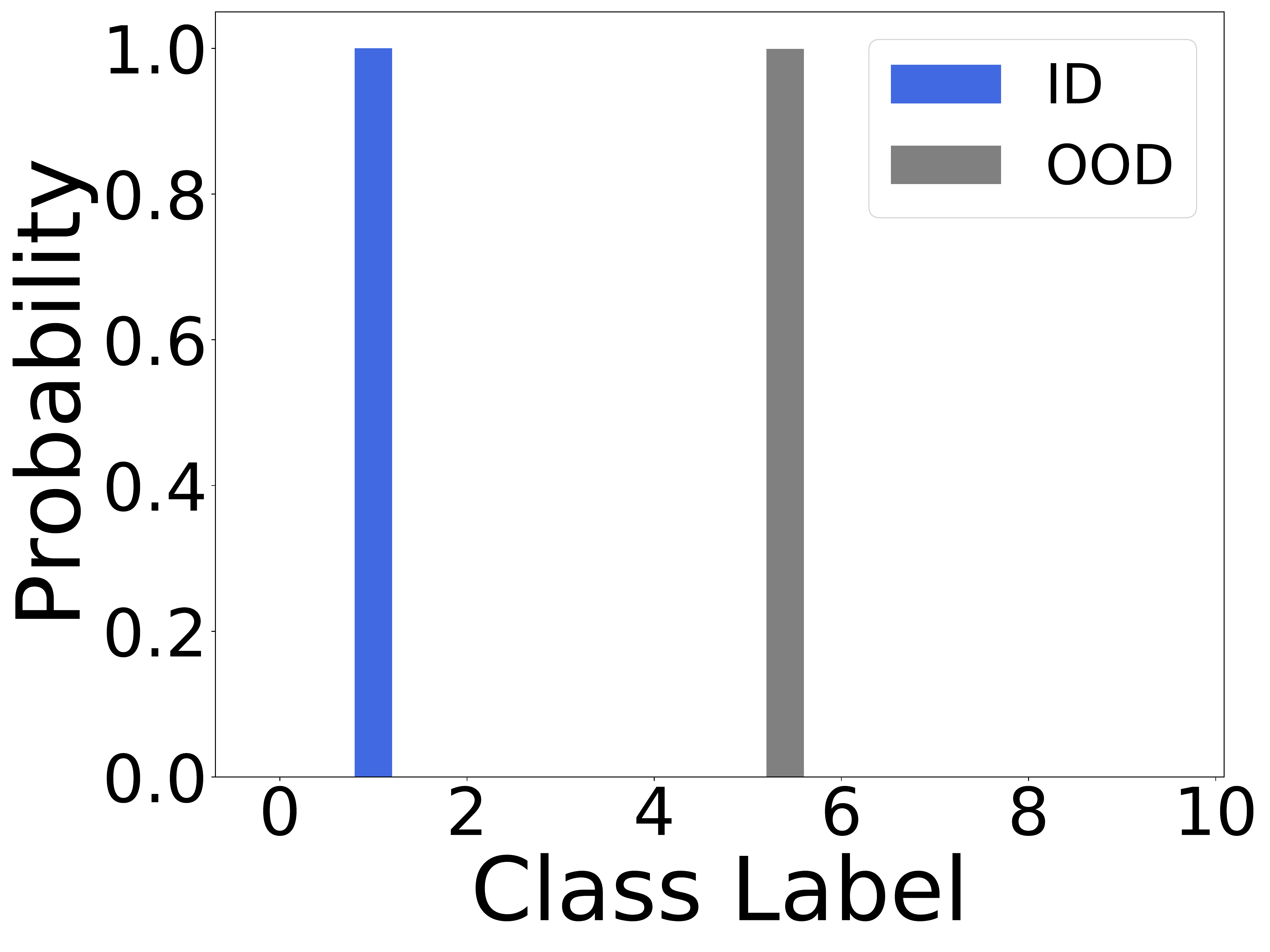}
        \caption{Cross-entropy Loss}
        \label{fig:msp_ce}
    \end{subfigure}
    \begin{subfigure}[b]{0.22\textwidth}
        \centering
        \includegraphics[height=3.0cm,width=3.8cm]{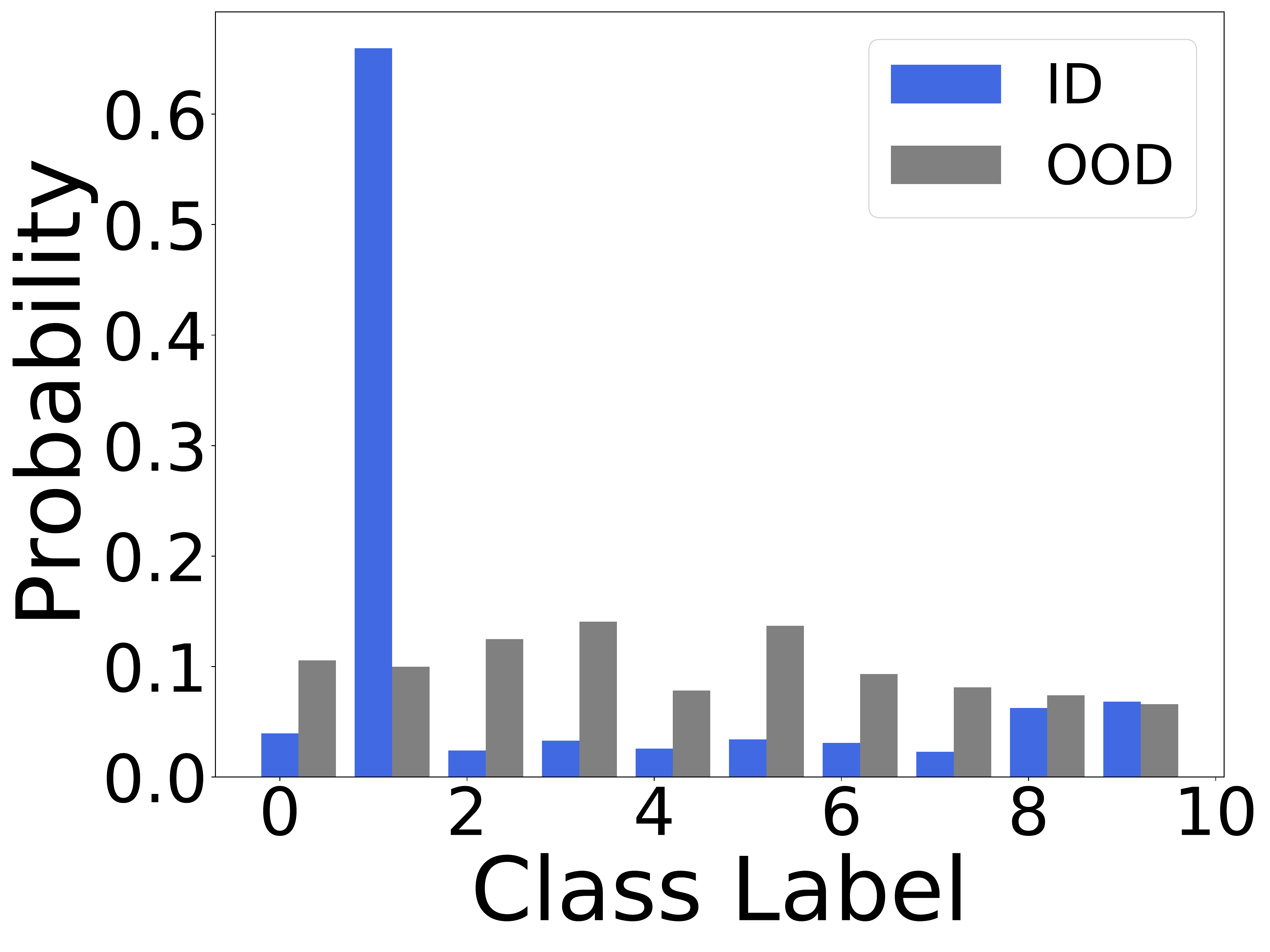}
        \caption{LogitNorm Loss (ours)}
        \label{fig:msp_logitnorm}
    \end{subfigure}
    \vspace{-0.3cm}
     \caption{The softmax outputs of two examples on a CIFAR-10 pre-trained WRN-40-2~\cite{zagoruyko2016wide} with (a) cross-entropy loss and (b) logit normalization loss. For the cross-entropy loss, the softmax confidence scores are 1.0 and 1.0 for the ID and OOD examples. In contrast, the softmax confidence scores of the network trained with LogitNorm loss are 0.66 and 0.14 for the ID and OOD examples. While using cross-entropy loss produces an extremely confident prediction for the OOD example, our method produces almost uniform softmax probabilities (near 0.1), which benefits OOD detection.
     }
     \label{fig:analysis}
    \vspace{-10pt}
\end{figure}

\begin{figure}[t]
    \centering
    \begin{subfigure}[b]{0.23\textwidth}
        \centering
        \includegraphics[height=3.0cm,width=4cm]{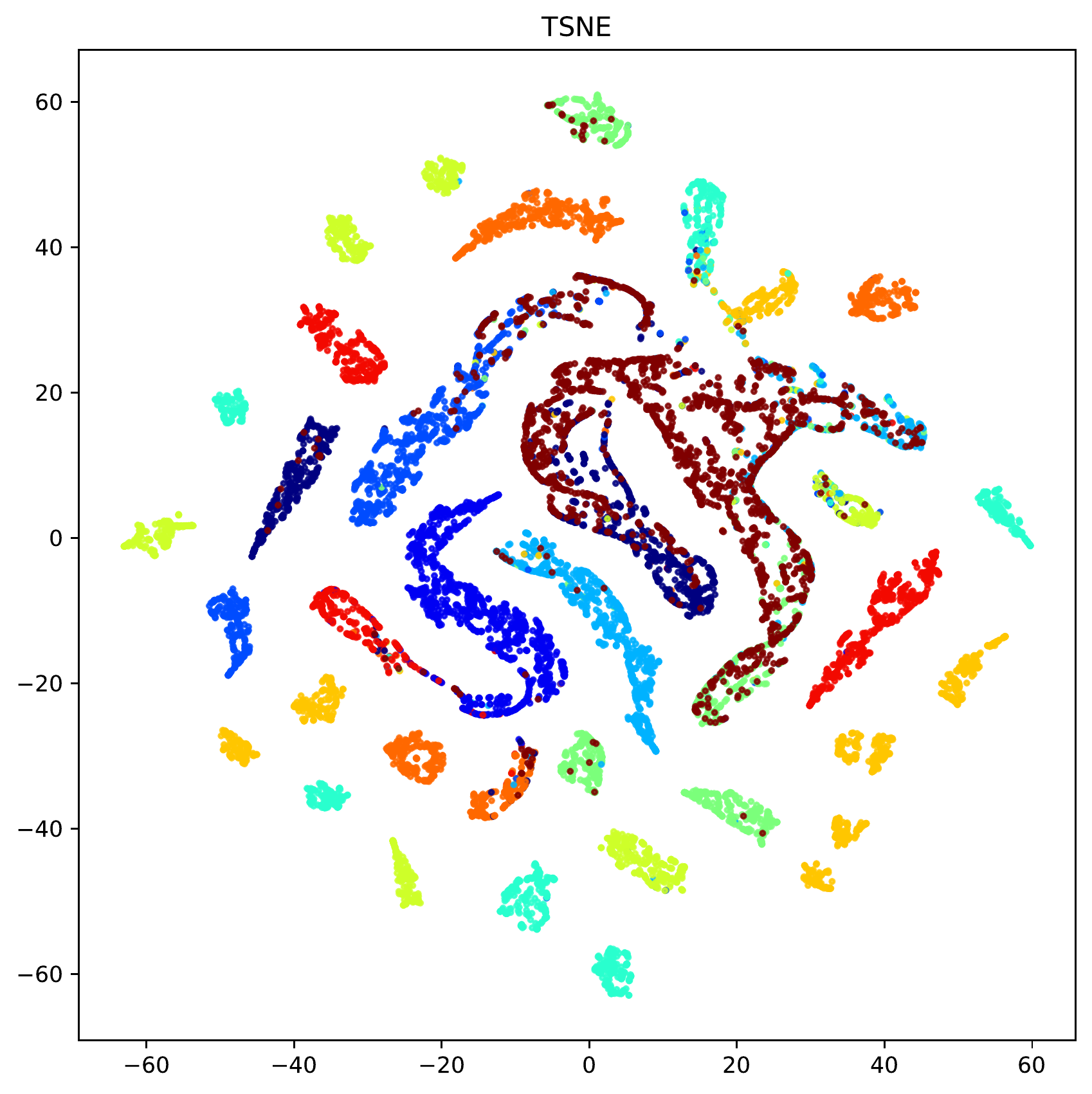}
        \caption{Cross-entropy Loss}
        \label{fig:msp_ce}
    \end{subfigure}
    \begin{subfigure}[b]{0.23\textwidth}
        \centering
        \includegraphics[height=3.0cm,width=4cm]{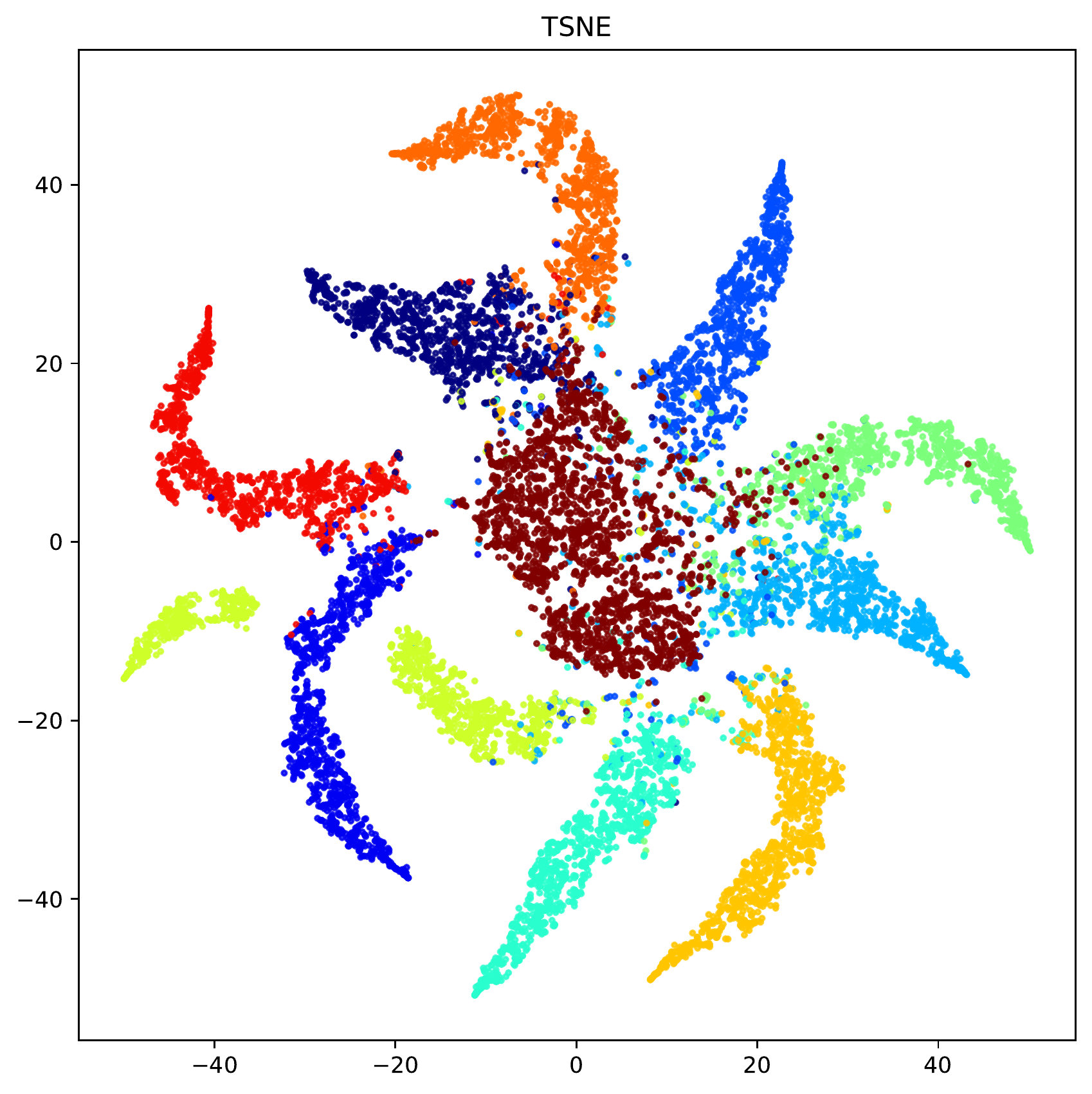}
        \caption{LogitNorm Loss (ours)}
        \label{fig:msp_logitnorm}
    \end{subfigure}
     \caption{t-SNE visualization \cite{van2008visualizing} of the softmax outputs from WRN-40-2~\cite{zagoruyko2016wide} trained on CIFAR-10 with (a) cross-entropy loss and (b) LogitNorm loss. All colors except for brown indicate 10 different ID classes. Points in \emph{brown} denote out-of-distribution examples from SVHN. Trained with logit normalization, the softmax outputs provide more meaningful information to differentiate in- and out-distribution samples.
     }
     \label{fig:tsne}
    \vspace{-15pt}
\end{figure}

By way of logit normalization, the magnitudes of the output vectors are strictly constant (\emph{i.e.}, $1/\tau$). Minimizing the loss in Eq.~(\ref{eq:norm_loss}) can only be achieved by adjusting the direction of the logit output $f$. The resulting model tends to give conservative predictions, especially for inputs that are far away from $\mathcal{P}_{\text{in}}$. We illustrate with an example in Figure~\ref{fig:analysis}, where training with logit normalization leads to softmax outputs that are more distinguishable between in- and out-of-distribution samples (right), as opposed to using cross-entropy loss (left). 
In Figure~\ref{fig:tsne}, we present a t-SNE visualization of the softmax outputs using Cross-entropy vs. LogitNorm Loss, where LogitNorm leads to more meaningful information to differentiate ID and OOD samples in the softmax output space. Below we further provide a lower bound of this new loss function in Eq.~(\ref{eq:norm_loss}).

\begin{proposition}[Lower Bound of Loss]
\label{prop:bound}
For any input $\boldsymbol{x}$ and any positive number $\tau \in \mathbb{R}^{+}$, the per-sample loss defined in Eq.~(\ref{eq:norm_loss}) has a lower bound: $ \mathcal{L}_{\text{logit\_norm}} \geq \log \left(1 + (k-1)e^{-2/\tau}\right)$,  where $k$ is the number of classes.
\end{proposition}

The proof of Proposition~\ref{prop:bound} is provided in Appendix~\ref{app:proofs_3}. From Proposition~\ref{prop:bound}, we find that the LogitNorm loss has a lower bound that depends on $\tau$ and number of classes $k$. In particular, it implies that the lower bound of the loss value increases with the value of $\tau$. For example, when $k=10$ and $\tau=1$, the norm of logits would be linearly scaled to 1 and the lower bound is about 0.7966. The high lower bound can cause optimization difficulty. For this reason, we found it is desirable to have a relatively small $\tau < 1$. We will analyze the effect of $\tau$ in detail in Section~\ref{sec:discussion}.

\begin{table*}[!t]
\centering
\caption{OOD detection performance comparison using softmax cross-entropy loss and LogitNorm loss. We use WRN-40-2 \cite{zagoruyko2016wide} to train on the in-distribution datasets and use softmax confidence score as the scoring function. All values are percentages. $\uparrow$ indicates larger values are better, and $\downarrow$ indicates smaller values are better. \textbf{Bold} numbers are superior results.}
\label{tab:cifar_results}
\renewcommand\arraystretch{1.2}
\resizebox{1.00\textwidth}{!}{
\setlength{\tabcolsep}{5mm}{
\begin{tabular}{c|ccc|ccc}
\toprule
 ID datasets  & \multicolumn{3}{c}{CIFAR-10} & \multicolumn{3}{c}{CIFAR-100} \\
\midrule
 OOD datasets  & FPR95 $\downarrow$ & AUROC $\uparrow$ & AUPR $\uparrow$ & FPR95 $\downarrow$ & AUROC $\uparrow$ & AUPR $\uparrow$\\
\midrule
 ~ &  \multicolumn{6}{c}{Cross-entropy loss / \textbf{LogitNorm loss (ours)}}\\
\midrule
 Texture &  64.13 / \textbf{28.64}  &  86.29 / \textbf{94.28}  &  96.28 / \textbf{98.63}   & 84.11 / \textbf{70.67}  &  74.05 / \textbf{78.65}  &  93.45 / \textbf{93.66}  \\
SVHN & 50.33 / \textbf{8.03}  &  93.48 / \textbf{98.47}  &  98.70 / \textbf{99.68}  &  79.09 / \textbf{45.98}  &  78.62 / \textbf{92.48}  &  94.99 / \textbf{98.45} \\
LSUN-C &  33.34 / \textbf{2.37}  &  95.29 / \textbf{99.42}  &  99.04 / \textbf{99.88}  &  67.94 / \textbf{13.93}  &  83.60 / \textbf{97.56}  &  96.32 / \textbf{99.48}  \\
LSUN-R &   42.52 / \textbf{10.93}  &  93.74 / \textbf{97.87}  &  98.66 / \textbf{99.59} &  82.21 / \textbf{68.68}  &  69.45 / \textbf{84.77}  &  91.45 / \textbf{96.52}   \\
iSUN & 46.56 / \textbf{12.28}  &  93.13 / \textbf{97.73}  &  98.55 / \textbf{99.56} &   84.50 / \textbf{71.47}  &  69.29 / \textbf{83.79}  &  91.44 / \textbf{96.24}  \\
Places365 &  60.23 / \textbf{31.64}  &  87.36 / \textbf{93.66}  &  96.62 / \textbf{98.49} &  81.09 / \textbf{80.20}  &  75.61 / \textbf{77.14}  &  93.74 / \textbf{94.16}  \\
\midrule
\textbf{average} &  49.52 / \textbf{15.65}  &  91.55 / \textbf{96.91}  &  97.98 / \textbf{99.31}  &  79.82 / \textbf{58.49}  &  75.10 / \textbf{85.73}  &  93.57 / \textbf{96.42}  \\
\bottomrule
\end{tabular}
}`
}
\end{table*}

\begin{figure*}
    \centering
    \begin{subfigure}[b]{0.48\textwidth}
        \centering
        \includegraphics[height=4.0cm,width=6cm]{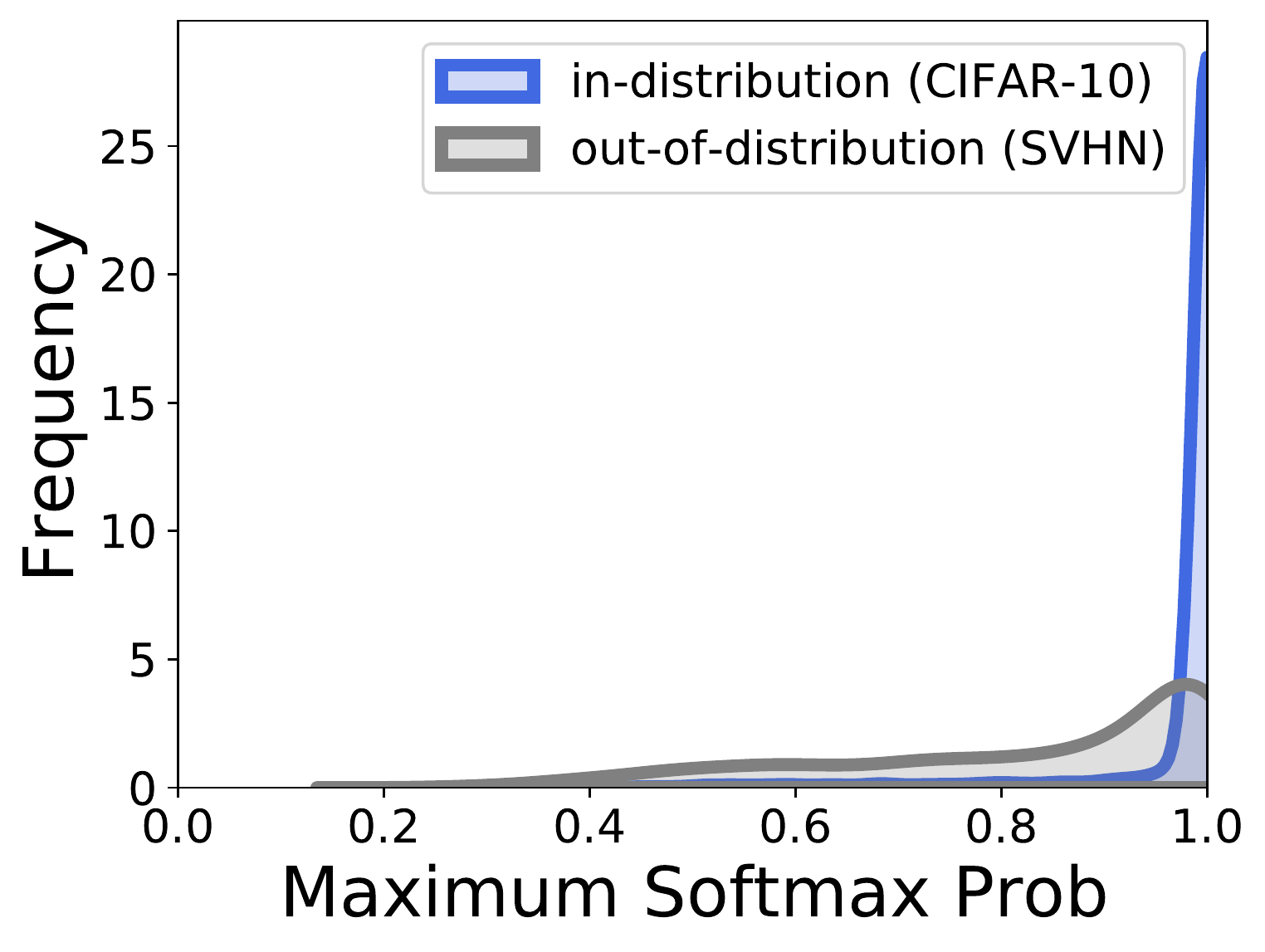}
        \caption{Cross-entropy Loss}
        \label{fig:pdf_ce}
    \end{subfigure}
    \begin{subfigure}[b]{0.48\textwidth}
        \centering
        \includegraphics[height=4.0cm,width=6cm]{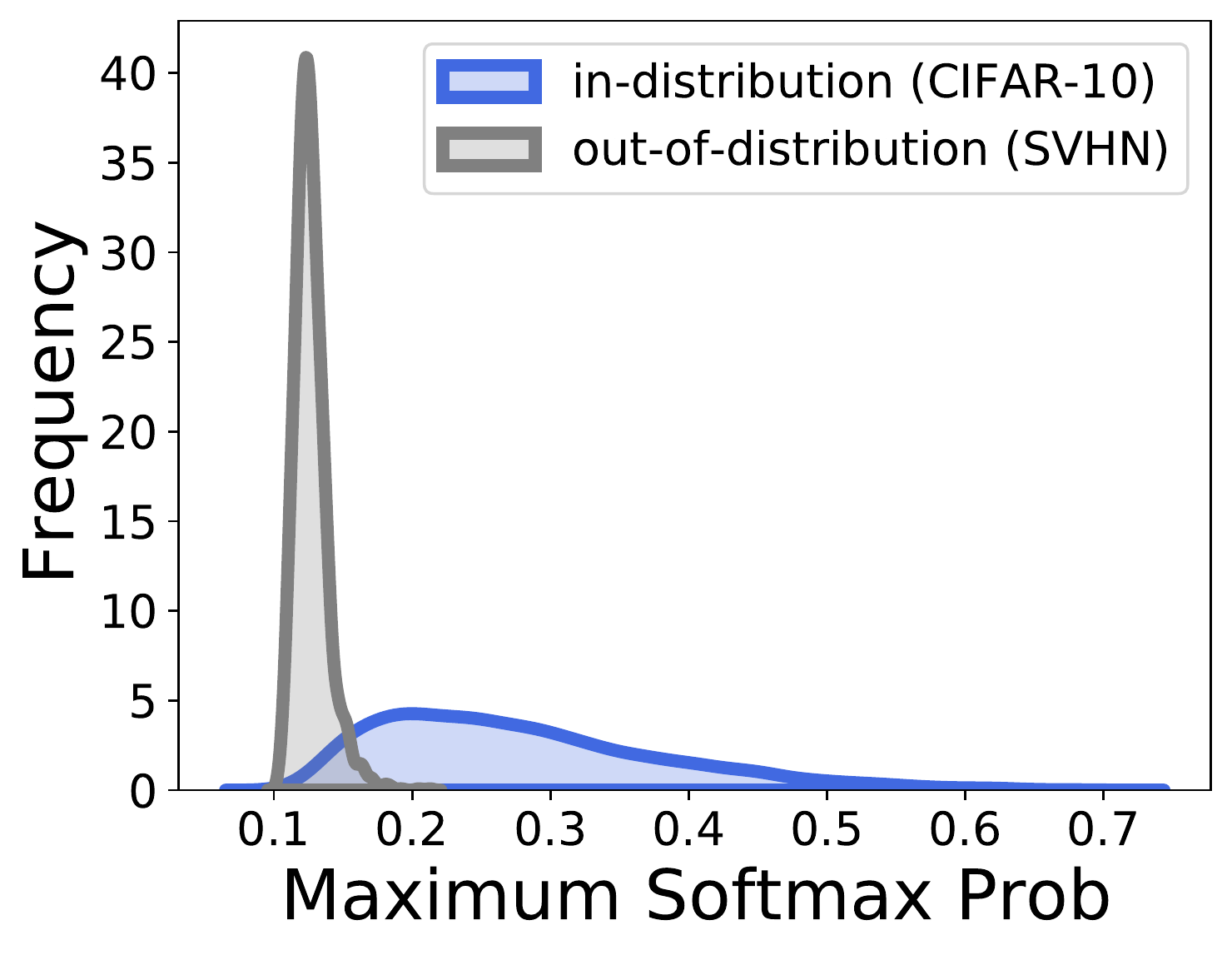}
        \caption{LogitNorm Loss (Ours)}
        \label{fig:pdf_logitnorm}
    \end{subfigure}
     \caption{Distribution of softmax confidence score from WRN-40-2~\cite{zagoruyko2016wide} trained on CIFAR-10 with (a) cross-entropy loss and (b) LogitNorm loss.}
     \label{fig:pdf}
    \vspace{-8pt}
\end{figure*}

\section{Experiments}

In this section, we verify the effectiveness of LogitNorm loss in OOD detection with several benchmark datasets.

\subsection{Experimental Setup}

\textbf{In-distribution datasets.} In this work, we use the CIFAR-10 and CIFAR-100~\cite{krizhevsky2009learning} datasets as in-distribution datasets, which are common benchmarks for OOD detection. Specifically, we use the standard split with 50,000 training images and 10,000 test images. 
All the images are of size $32 \times 32$.

\textbf{Out-of-distribution datasets.} For the OOD detection evaluation, we use six common benchmarks as OOD test datasets $\mathcal{D}_{\text{out}}^{\text{test}}$: \texttt{Textures} \cite{cimpoi2014describing}, \texttt{SVHN} \cite{netzer2011reading}, \texttt{Places365} \cite{zhou2017places}, \texttt{LSUN-Crop} \cite{yu2015lsun}, \texttt{LSUN-Resize} \cite{yu2015lsun}, and \texttt{iSUN} \cite{xu2015turkergaze}. For all test datasets, the images are of size $32 \times 32$. The detail information of the six datasets is presented in Appendix~\ref{app:ood_datasets}.

\textbf{Evaluation metrics.} We evaluate the performance of OOD detection by measuring the following metrics: (1) the false positive rate (FPR95) of OOD examples when the true positive rate of in-distribution examples is 95\%; (2) the area under the receiver operating characteristic curve (AUROC); and (3) the area under the precision-call curve (AUPR).

\textbf{Training details.} For main results, we perform training with WRN-40-2 \cite{zagoruyko2016wide} on CIFAR-10/100. The network is trained for 200 epochs using SGD with a momentum of 0.9, a weight decay of 0.0005, a dropout rate of 0.3, and a batch size of 128. We set the initial learning rate as 0.1 and reduce it by a factor of 10 at 80 and 140 epochs. The hyperparameter $\tau$ is selected from the range $\{ 0.001, 0.005, 0.01, \ldots, 0.05\}$. We set $0.04$ for CIFAR-10 by default. For hyperparameter tuning, we use Gaussian noises as the validation set. All experiments are repeated five times with different seeds, and we report the average performance. We conduct all the experiments on NVIDIA GeForce RTX 3090 and implement all methods with default parameters using PyTorch \cite{NEURIPS2019_9015}.

\begin{table*}[t]
\centering
\caption{OOD detection performance comparison with various scoring functions using softmax cross-entropy loss and logit normalization loss. We use WRN-40-2 to train on the in-distribution datasets. All values are percentages. $\uparrow$ indicates larger values are better, and $\downarrow$ indicates smaller values are better. \textbf{Bold} numbers are superior results.}
\label{tab:other_score}
\renewcommand\arraystretch{1.1}
\resizebox{1.00\textwidth}{!}{
\setlength{\tabcolsep}{5mm}{
\begin{tabular}{c|ccc|ccc}
\toprule
ID datasets  & \multicolumn{3}{c}{CIFAR-10} & \multicolumn{3}{c}{CIFAR-100} \\
\midrule
Score  &FPR95 $\downarrow$ & AUROC $\uparrow$ & AUPR $\uparrow$ &FPR95 $\downarrow$ & AUROC $\uparrow$ & AUPR $\uparrow$\\
\midrule
 ~ &  \multicolumn{6}{c}{Cross-entropy loss / \textbf{LogitNorm loss (ours)}}\\
\midrule
Softmax   &  49.52 / \textbf{15.65}  &  91.55 / \textbf{96.91}  &  97.98 / \textbf{99.31}  &  79.82 / \textbf{58.35}  &  75.10 / \textbf{85.75}  &  93.57 / \textbf{96.44}  \\
ODIN   &  40.32 / \textbf{12.95}  &  87.21 / \textbf{97.37}  &  96.24 / \textbf{99.40}  &  70.71 / \textbf{58.13}  &  81.38 / \textbf{85.65}  &  95.37 / \textbf{96.36}  \\
Energy    & 26.82 / \textbf{19.14}  &  93.07 / \textbf{96.09}  &  98.02 / \textbf{99.14} &  70.87 / \textbf{65.46}  &  81.45 / \textbf{84.84}  &  95.38 / \textbf{96.27} \\
GradNorm    & 58.98 / \textbf{17.78}  &  72.29 / \textbf{96.34}  &  90.75 / \textbf{99.14} &  87.01 / \textbf{61.89}  &  52.84 / \textbf{81.41}  &  84.12 / \textbf{94.85} \\
\bottomrule
\end{tabular}
}
}
\vspace{-15pt}
\end{table*}

\subsection{Results}
\label{sec:results}

\begin{table}[ht]
\centering
\caption{OOD detection performance comparison trained on CIFAR-10 with different network architectures: WRN-40-2 \cite{zagoruyko2016wide}, ResNet-34 \cite{he2016deep}, DenseNet-BC \cite{huang2017densely}. All values are percentages. $\uparrow$ indicates larger values are better, and $\downarrow$ indicates smaller values are better. \textbf{Bold} numbers are superior results.}
\label{tab:arch}
\renewcommand\arraystretch{1.4}
\resizebox{0.50\textwidth}{!}{
\setlength{\tabcolsep}{1mm}{
\begin{tabular}{c|cccc}
\toprule
Architecture  & ID Accuracy $\uparrow$  &FPR95 $\downarrow$ & AUROC $\uparrow$ & AUPR $\uparrow$ \\
\midrule
 ~ &  \multicolumn{4}{c}{Cross-entropy loss / \textbf{LogitNorm loss (ours)}}\\
\midrule
WRN-40-2   & 94.75 / 94.69 & 49.52 / \textbf{15.65}  &  91.55 / \textbf{96.91}  &  97.98 / \textbf{99.31}   \\
ResNet-34  & 95.01 / 95.14  & 47.74 / \textbf{15.82}  &  91.15 / \textbf{97.01}  &  97.72 / \textbf{99.32}   \\
DenseNet   & 94.55 / 94.37 & 50.41 / \textbf{18.57}  &  91.48 / \textbf{96.16}  &  98.04 / \textbf{99.10}  \\
\bottomrule
\end{tabular}
}
}
\vspace{-10pt}
\end{table}

\textbf{How does logit normalization influence OOD detection performance?} In Table~\ref{tab:cifar_results}, we compare the OOD detection performance on models trained with cross-entropy loss and LogitNorm loss respectively. To isolate the effect of the loss function in training, we keep the test-time OOD scoring function to be the same, \emph{i.e.}, softmax confidence score: 
$$
S(\boldsymbol{x}) = \max_i \frac{e^{f_i(\boldsymbol{x}; {\theta})}}{\sum^{k}_{j=1} e^{f_{j}(\boldsymbol{x}; {\theta})}}.
$$
A salient observation is that our method drastically improves the OOD detection performance by employing LogitNorm loss. For example, on the CIFAR-10 model, when evaluated against SVHN as OOD data, our method reduces the FPR95 from 50.33\% to 8.03\%---a \textbf{42.3}\% of direct improvement. Averaged across six test datasets, our approach reduces the FPR95 by \textbf{33.87}\% compared with using MSP on the model trained with cross-entropy loss. On CIFAR-100, our method also improves the performance by a meaningful margin.

To further illustrate the difference between the two losses on OOD detection, we visualize and compare the distribution of softmax confidence score for ID and OOD data, derived from networks trained with cross-entropy vs. LogitNorm losses. With cross-entropy loss, the softmax scores for both ID and OOD data concentrate on high values, as shown in Figure~\ref{fig:pdf_ce}. In contrast, the network trained with LogitNorm loss produces highly distinguishable scores between ID and OOD data. From Figure~\ref{fig:pdf_logitnorm}, we observe that the softmax confidence score for most OOD examples is around 0.1, which indicates that the softmax outputs are close to a uniform distribution. Overall the experiments show that training with LogitNorm loss makes the softmax scores more distinguishable between in- and out-of-distributions and consequently enables more effective OOD detection.

\textbf{Can the logit normalization improve existing scoring functions?} In Table~\ref{tab:other_score}, we show that the LogitNorm loss not only outperforms, but also boosts competitive OOD scoring functions. Note that all the OOD scoring functions considered are originally developed based on models trained with cross-entropy loss, hence are natural choices of comparison. In particular, we consider: 1) \textbf{MSP} \cite{hendrycks2016baseline} uses the softmax confidence score to detect OOD samples. 2) \textbf{ODIN} \cite{liang2018enhancing} employs temperature scaling and input perturbation to improve OOD detection. Following the original setting, we use the temperature parameter $T=1000$ and $\epsilon = 0.0014$. 
3) \textbf{Energy score} \cite{liu2020energy} utilizes the information in logits for OOD detection, which is the negative log of the denominator in softmax function: $E(\mathbf{x} ; f)=-T \cdot \log \sum_{i=1}^{k} e^{f_{i}(\boldsymbol{x}) / T}$. With LogitNorm loss, we set $T = 0.1$ for CIFAR-10 and $T=0.01$ for CIFAR-100. 4) \textbf{GradNorm} \cite{huang2021importance} detects OOD inputs by utilizing information extracted from the gradient space.

Our results in Table~\ref{tab:other_score} suggest that logit normalization can benefit a wide range of downstream OOD scoring functions. Due to space constraints, we report the average performance across six test OOD datasets. The OOD detection performance on each OOD test dataset is provided in Appendix~\ref{app:detail_results}. For example, we observe that the FPR95 of the ODIN method is reduced from 40.32\% to 12.95\% when employing logit normalization, establishing strong performance. In addition, we find that logit normalization enables the energy score and GradNorm score to achieve decent OOD detection performance as well.

\textbf{Logit normalization is effective on different architectures.} In Table \ref{tab:arch}, we show that LogitNorm is effective on a diverse range of model architectures. The results are based on softmax confidence score as test-time OOD score. In particular, our method consistently improves the performance using WRN-40-2~\cite{zagoruyko2016wide}, ResNet~\cite{he2016deep} and DenseNet~\cite{huang2017densely} architectures. For example, on DenseNet, using LogitNorm loss reduces the average FPR95 from 50.41\% to 18.57\%. 

\begin{table}[!t]
\centering
\caption{Comparison results of ECE ($\%$) with $M=15$, using WRN-40-2 trained on CIFAR-10. For temperature scaling (TS), $T$ is tuned on a hold-out validation set by optimization methods.}
\label{tab:calibration}
\renewcommand\arraystretch{1.4}
\resizebox{0.50\textwidth}{!}{
\setlength{\tabcolsep}{5mm}{
\begin{tabular}{c|ccc}
\toprule
Dataset    & &Cross Entropy & LogitNorm (ours)  \\
\midrule
\multirow{2}*{CIFAR-10} & w/o TS & 3.20  &  66.95   \\
\cline{2-4}
~  & w/ TS & 0.66  &  \textbf{0.41}  \\
\midrule
\multirow{2}*{CIFAR-100} & w/o TS & 11.69  &  70.45   \\
\cline{2-4}
~  & w/ TS & 2.18 &  \textbf{1.67} \\
\bottomrule
\end{tabular}
}
}
\vspace{-20pt}
\end{table}

\textbf{Logit normalization maintains classification accuracy.} We also verify whether LogitNorm loss affects the classification accuracy. Our results in Table~\ref{tab:arch} show that LogitNorm can improve the OOD detection performance, and at the same time, achieve similar accuracy as the cross-entropy loss. For example, when trained on ResNet-34 with CIFAR-10 as the ID dataset, using LogitNorm loss achieves a test accuracy of $95.14\%$ on CIFAR-10, on par with the test accuracy  $95.01\%$ using cross-entropy loss. On CIFAR-100, LogitNorm loss and cross-entropy loss also achieves comparable test accuracies ($75.12\%$ vs. $75.23\%$). Overall LogitNorm loss maintains comparable classification accuracy on the ID data while leading to significant improvement in OOD detection performance. 

\textbf{Logit normalization enables better calibration.} While the OOD detection task concentrates on the separability between ID and OOD data, the calibration task focuses solely on the ID data---softmax confidence score should represent the true probability of correctness~\cite{guo2017calibration}. In practice, Expected Calibration Error (ECE) \cite{naeini2015obtaining} is commonly used to measure the calibration performance from finite samples. In Figure~\ref{fig:pdf}, we observe that LogitNorm loss leads to a smoother distribution of softmax confidence score for ID examples, in contrast to the spiky distribution induced by cross-entropy loss (\emph{i.e.}, values concentrate around 1). 
It implies that the model trained with LogitNorm loss preserves distinguishable information for different ID samples, indicating its potential in improving model calibration. Indeed, our results in Table~\ref{tab:calibration} show that the model trained with LogitNorm achieves better calibration performance by way of post-hoc temperature scaling.

\section{Discussion}
\label{sec:discussion}

\textbf{Logit normalization vs. Logit penalty.} While our logit normalization has demonstrated strong promise, a question arises: \emph{can a similar effect be achieved by imposing a penalty on the $L_2$ norm of the logits}? In this ablation, we show that explicitly constraining logit norm via a Lagrangian multiplier~\cite{forst2010optimization} does not work well. Specifically, we consider the alternative loss:
\begin{equation*}
\mathcal{L}_{\text{logit\_penalty}}(f(\boldsymbol{x};{\theta}),y) = 
\mathcal{L}_{\text{CE}}(f(\boldsymbol{x};{\theta}),y) + \lambda {\|f(\boldsymbol{x};{\theta})\|}_2.
\end{equation*}
where $\lambda$ denotes the Lagrangian multiplier that controls the trade-off between the cross-entropy loss and the regularization term.

Our results in Table \ref{tab:penalty} show that both logit penalty and logit normalization lead to logits with small $L_2$ norms, compared with using cross-entropy loss. However, unlike LogitNorm, the logit penalty method produces a large $L_2$ norm for OOD data as well, resulting in the inferior performance of OOD detection. 
In practice, we notice that the network trained with the logit penalty can suffer from optimization difficulty and sometimes fail to converge if $\lambda$ is too large (which is needed to regularize the logit norm effectively). Overall, we show that simply constraining the logit norm during training cannot help the OOD detection task while our LogitNorm loss significantly improves the performance.

\begin{table}[!t]
\centering
\caption{OOD detection performance comparison with different loss functions. We use WRN-40-2 trained on CIFAR-10. \textbf{Bold} numbers are superior results. We report the average norm value of logits as $L_2$ Norm, where we use SVHN as OOD test dataset.}
\label{tab:penalty}
\renewcommand\arraystretch{1.2}
\resizebox{0.50\textwidth}{!}{
\setlength{\tabcolsep}{2mm}{
\begin{tabular}{c|cccc}
\toprule
Loss    &FPR95 $\downarrow$ & AUROC $\uparrow$ & AUPR $\uparrow$ &  $L_2$ Norm (ID / OOD) \\
\midrule
CrossEntropy  & 49.52  &  91.55  &  97.98  &  12.80 / 10.91   \\
LogitPenalty    & 57.62  &  73.27  &  87.62  & 1.90 / 1.74 \\
LogitNorm (ours)   & \textbf{15.65}  &  \textbf{96.91}  &  \textbf{99.31}  &  1.48 / 0.49\\
\bottomrule
\end{tabular}
}
}
\vspace{-10pt}
\end{table}

\begin{table}[t]
\centering
\caption{Comparison between GODIN and LogitNorm. We use WRN-40-2 \cite{zagoruyko2016wide} trained on CIFAR-10. For a fair comparison, we use the ODIN score for LogitNorm loss. \textbf{Bold} numbers are superior results.}
\label{tab:GODIN}
\renewcommand\arraystretch{1.2}
\resizebox{0.45\textwidth}{!}{
\setlength{\tabcolsep}{4mm}{
\begin{tabular}{c|cccc}
\toprule
Loss    &FPR95 $\downarrow$ & AUROC $\uparrow$ & AUPR $\uparrow$ \\
\midrule
GODIN  &  25.24 &  95.07  &  98.93     \\
LogitNorm (ours) &  \textbf{15.65}  &  \textbf{96.91}  &  \textbf{99.31}\\
\bottomrule
\end{tabular}
}
}
\vspace{-10pt}
\end{table}

\textbf{Relations to temperature scaling.} As introduced in Section~\ref{sec:method}, the logit normalization can be viewed as an \emph{input-dependent} temperature on the logits. 
Related to our work, previous work ODIN \cite{liang2018enhancing} proposes a variant of softmax score by employing  temperature scaling with a \emph{constant} $T$ in the testing phase:
$$
S(\boldsymbol{x}) = \max_i \frac{e^{f_i(\boldsymbol{x}; {\theta})/T}}{\sum^{k}_{j=1} e^{f_{j}(\boldsymbol{x}; {\theta})/T}},
$$
where the temperature is the same for all inputs. In contrast, our method bares two key differences: (1) the effective temperature in LogitNorm is input-dependent rather than a global constant, and (2) the temperature in LogitNorm can be enforced during the training stage. Our method is compatible with test-time temperature scaling in OOD detection and can improve calibration performance with temperature scaling, as we show in Section~\ref{sec:results}.

\begin{figure}[!t]
    \centering
    \includegraphics[width=0.30\textwidth]{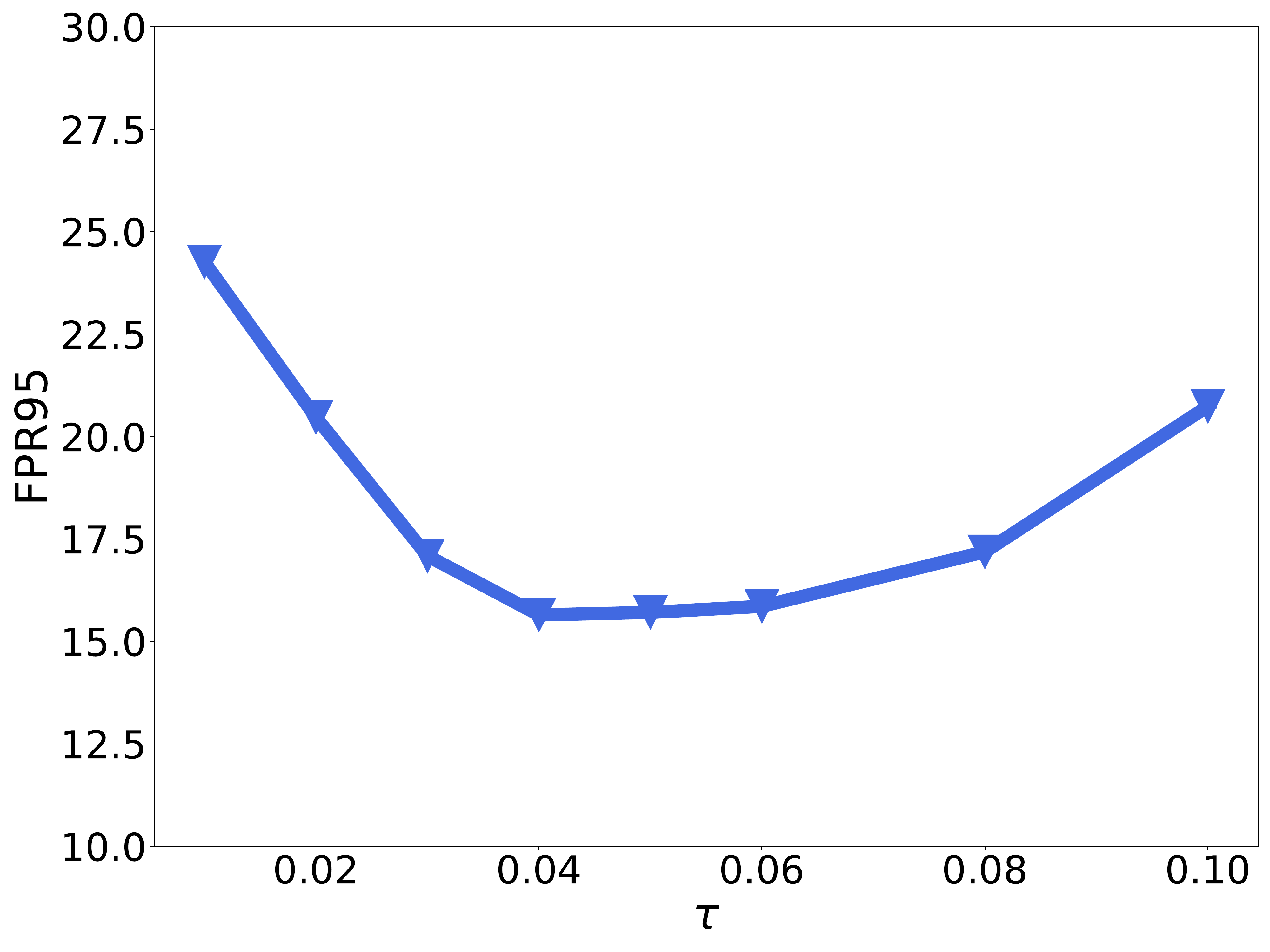}
    \vspace{-0.3cm}
    \caption{Effect of $\tau$ in LogitNorm with CIFAR-10.}
    \label{fig:tau}
    \vspace{-5pt}
\end{figure}

In Figure~\ref{fig:tau}, we further ablate how the parameter $\tau$ in our method (\emph{cf.} Eq.~\ref{eq:norm_loss}) affects the OOD detection performance. The analysis is based on CIFAR-10, with FPR95 averaged across six test datasets. Our results echo the analysis in  Proposition~\ref{prop:bound}, where a large $\tau$ would lead to a large lower bound on the loss, which is less desirable from the optimization perspective.

\textbf{Relations to other normalization methods. } 
In the literature, normalization method is applied for OOD detection in the form of cosine similarity \cite{techapanurak2020hyperparameter,hsu2020generalized}.
In particular, Cosine loss \cite{techapanurak2020hyperparameter} and Generalized ODIN (GODIN) \cite{hsu2020generalized} decompose the logit $f_i(\mathbf{x}; \mathbf{\theta})$ for class $i$ as:
$
f_{i}(\mathbf{x}; \mathbf{\theta})=\frac{h_{i}(\mathbf{x}; \mathbf{\theta})}{g(\mathbf{x}; \mathbf{\theta})}
$, where
$$
\begin{cases}
h_{i}(\boldsymbol{x})= \cos(\boldsymbol{w}_i, \phi^{p}(\boldsymbol{x})) = \frac{\boldsymbol{w}_{i}^{T} \phi^{p}(\boldsymbol{x})}{\left\|\boldsymbol{w}_{i}\right\|\left\|\phi^{p}(\boldsymbol{x})\right\|},\\
g(\mathbf{x}; \mathbf{\theta}) = \exp \left\{\mathrm{BN}\left(\boldsymbol{W}^{\top}\phi^p(\boldsymbol{x})+b \right)\right\}.
\end{cases}
$$
In the testing phase, the maximum cosine similarity is used as the scoring function. Contrastingly, LogitNorm has two key differences: (1) the cosine similarity in $h_i(\boldsymbol{x})$ applies $L_2$ normalization on the last-layer weight $\boldsymbol{w}$ and the learned feature $\phi^{p}(\boldsymbol{x})$, while our LogitNorm normalizes the network output $f(\boldsymbol{x};\theta)$; (2) our LogitNorm can boost the performance of common OOD scoring functions, while GODIN and Cosine loss detect OOD examples with their specific scoring functions. In Table~\ref{tab:GODIN}, we show our LogitNorm with the MSP score achieves better performance than GODIN in OOD detection. More discussion on future work is provided in Appendix.~\ref{app:limit}.

\section{Related Work}
\label{sec:relate}

\textbf{OOD detection.} OOD detection is an increasingly important topic for deploying machine learning models in the open world and has attracted a surge of interest in two directions. 

1) Some methods aim to design scoring functions for OOD detection, such as OpenMax score \cite{bendale2016towards}, maximum softmax probability \cite{hendrycks2016baseline}, ODIN score \cite{liang2018enhancing, hsu2020generalized}, Mahalanobis distance-based score \cite{lee2018simple}, Energy-based score \cite{liu2020energy, wang2021can, morteza2022provable}, ReAct \cite{sun2021react}, GradNorm score \cite{huang2021importance}, and non-parametric KNN-based score~\cite{sun2022knn, zhu@detect}. In this work, we first show that logit normalization can drastically mitigate the overconfidence issue for OOD data, thereby boosting the performance of existing scoring functions in OOD detection.

2) Some works address the out-of-distribution detection problem by training-time regularization \cite{lee2017training,  bevandic2018discriminative,hendrycks2019oe, geifman2019selectivenet, malinin2018predictive, mohseni2020self, jeong2020ood, liu2020energy, chen2021atom, wei2021odnl, wei@open, ming2022posterior}. For example, models are encouraged to give predictions with uniform distribution~\cite{lee2017training, hendrycks2019oe} or higher energies~\cite{liu2020energy,du2022vos, ming2022posterior, du2022unknown, katzsamuels2022training} for outliers. The energy-based regularization has a direct theoretical interpretation as shaping the log-likelihood, hence naturally suits OOD detection. Contrastive learning methods are also employed for the OOD detection task \cite{tack2020csi,sehwag2021ssd, ming2022cider}, which 
can be computationally more expensive to train than ours. In this work, we focus on exploring classification-based loss functions for OOD detection, which only requires in-distribution data in training. LogitNorm is easy to implement and use, and maintains the same training scheme as standard cross-entropy loss. 

\textbf{Normalization in deep learning.} In the literature, normalization has been widely used in metric learning \cite{Sohn2016ImprovedDM, wu2018unsupervised, Oord2018RepresentationLW}, face recognition \cite{ranjan2017l2, liu2017sphereface, wang2017normface, wang2018cosface, deng2019arcface, zhang2019adacos}, and self-supervised learning \cite{chen2020simple}. $L_2$-constrained softmax \cite{ranjan2017l2} applies the $L_2$ normalization on features and SphereFace \cite{liu2017sphereface} normalizes the weights of the last inner-product later only. 
Cosine loss \cite{wang2017normface, wang2018cosface} normalizes both the features and weights to achieve better performance for face verification. LayerNorm \cite{xu2019understanding} normalizes the distributions of intermediate layers for better generalization accuracy. In self-supervised learning, SimCLR \cite{chen2020simple} adopts cosine similarity to measure the feature distances between positive pair of examples. A recent study \cite{kornblith2021why} shows that several loss functions, including logit normalization and cosine softmax, lead to higher accuracy on ImageNet but degrade the performance of transfer tasks. 
In addition, GODIN \cite{hsu2020generalized} and Cosine loss \cite{techapanurak2020hyperparameter} adopt cosine similarity for better performance on OOD detection. As discussed in Section~\ref{sec:discussion}, our method is superior to these cosine-based methods, since it is applicable to existing scoring functions and achieves strong performance in OOD detection.

\textbf{Confidence calibration.} Confidence calibration has been studied in various contexts in recent years. Some works address the miscalibration problem by post-hoc methods, such as Temperature Scaling \cite{platt1999probabilistic, guo2017calibration} and Histogram Binning \cite{zadrozny2001obtaining}. Besides, some regularization methods are also proposed to improve the calibration quality of deep neural networks, like weight decay \cite{guo2017calibration}, label smoothing \cite{Szegedy2016rethinking,Mller2019WhenDL}, and focal loss \cite{Lin2017FocalLF, mukhoti2020calibrating}. Conformal prediction based method \cite{lei2013distribution} outputs the empty set as prediction in case of too high ``atypicality''. Top-label calibration aims to calibrate the reported probability for the predicted class label \cite{gupta2022toplabel}. Recent work \cite{neurips21dbwang} shows that these regularization methods make it harder to further improve the calibration performance with post-hoc methods. LogitNorm loss yields better calibration performance with Temperature Scaling than cross-entropy.

\section{Conclusion}

In this paper, we introduce Logit Normalization (LogitNorm), a simple alternative to the cross-entropy loss that enhances many existing post-hoc methods for OOD detection. 
By decoupling the influence of logits' norm from the training objective and its optimization, the model tends to give conservative predictions for OOD inputs, resulting in a stronger separability from ID data. Extensive experiments show that LogitNorm can improve both OOD detection and confidence calibration while maintaining the classification accuracy on ID data. This method can be easily adopted in practical settings. It is straightforward to implement with existing deep learning frameworks, and does not require sophisticated changes to the loss or training scheme. We hope that our insights inspire future research to further explore loss function design for OOD detection.


\section*{Acknowledgements}

This research is supported by MOE Tier-1 project RG13/19 (S). LF is supported by the National Natural Science Foundation of China (Grant No. 62106028). YL is supported by Wisconsin Alumni Research Foundation (WARF), Facebook Research Award, and a Google-Initiated Focused Research Award.

\bibliography{main}
\bibliographystyle{icml2022}

\clearpage  
\appendix

\onecolumn
\section{Proof of Proposition~\ref{prop:magnitude}}
\label{app:proofs_1}

From Eq.~(\ref{eq:decouple}), we have $\boldsymbol{f} = \|\boldsymbol{f}\| \cdot \boldsymbol{\hat{f}}$.
Then,
$$\arg \max_{i}(f_i) = \arg \max_{i}(\lVert\boldsymbol{f}\rVert \cdot \hat{f}_i)
= \arg \max_{i}(\hat{f}_i).$$
Similarly, for any given scalar $s > 1$, we have,
$$
\arg \max_{i}(sf_i) = \arg \max_{i}(s\lVert\boldsymbol{f}\rVert \cdot \hat{f}_i)
= \arg \max_{i}(\hat{f}_i).
$$
Thus Proposition~\ref{prop:magnitude} is proved.\qed

\section{Proof of Proposition~\ref{prop:scale}}
\label{app:proofs_2}

From Proposition \ref{prop:magnitude}, we have $$\arg\max_i(f_i) = \arg\max_i(sf_i) = c.$$

Let $f_c = \max_i(f_i)$, and $t = s - 1$, then we have,
\begin{align*}
\sigma_c(s\boldsymbol{f}) &= \frac{e^{(1+t)f_c}}{\sum^n_{j=1} e^{(1+t)f_j}} \\
&= \frac{e^{f_c}}{\sum^n_{j=1} e^{f_j + t(f_j - f_c)}}.
\end{align*}
For any $j \in [1,2,\ldots,n]$, we have, $f_j - f_c \leq 0$. Then
$$
\sigma_c(s\boldsymbol{f}) \geq \frac{e^{f_c}}{\sum^n_{j=1} e^{f_j}} = \sigma_c(\boldsymbol{f}).
$$

Thus Proposition \ref{prop:scale} is proved.\qed

\section{Proof of Proposition~\ref{prop:bound}}
\label{app:proofs_3}

Let $\boldsymbol{\widetilde{f}} = \boldsymbol{f}/(\tau \|\boldsymbol{f}\|)$, then we have
$\|\boldsymbol{\widetilde{f}}\| = 1/\tau.$

That is, $\sum_{i=1}^k {\boldsymbol{\widetilde{f}}_i}^2 = {\|\boldsymbol{\widetilde{f}}\|}^2 = 1/\tau^2.$

Hence, $$-1/\tau \leq \boldsymbol{\widetilde{f}}_i \leq 1/\tau, \forall\ i \in 1, \ldots, k.$$
Let $\sigma(\boldsymbol{\widetilde{f}}) = \frac{e^{\boldsymbol{\widetilde{f}}_y}}{\sum^{k}_{i=1} e^{\boldsymbol{\widetilde{f}}_{i}}}$, then we have
\begin{align*}
\sigma(\boldsymbol{\widetilde{f}}) &\leq \frac{e^{1/\tau}}{e^{1/\tau} + (k-1) e^{-1/\tau}} \\
&= \frac{1}{1 + (k-1) e^{-2/\tau}}
\end{align*}
Hence,
\begin{align*}
\mathcal{L}_{\text{logit\_norm}} &= -\log(\sigma(\boldsymbol{\widetilde{f}})) \\
&\geq -\log \frac{1}{1 + (k-1) e^{-2/\tau}} \\
&= \log (1 + (k-1) e^{-2/\tau})
\end{align*}
Thus Proposition~\ref{prop:bound} is proved.\qed


\section{Descriptions of OOD Datasets}
\label{app:ood_datasets}

Following the prior literature, we use six OOD test datasets: \textit{Textures} \cite{cimpoi2014describing} is a dataset of describable textural images. 
\textit{SVHN} dataset \cite{netzer2011reading} contains 32 × 32 color images of house numbers, which has ten classes comprised of the digits 0-9.
\textit{LSUN} \cite{yu2015lsun} is another scene understanding dataset with fewer classes than Places365. Here we use \textit{LSUN-C} and \textit{LSUN-R} to denote the cropped and resized version of the LSUN dataset respectively.
\textit{iSUN} \cite{xu2015turkergaze} is a large-scale eye tracking dataset, selected from natural scene images of the SUN database \cite{xiao2010sun}. 
\textit{Places365} \cite{zhou2017places} consists in images for scene recognition rather than object recognition.

\section{Future Work}
\label{app:limit}

In this paper, we introduce a simple fix to the cross-entropy loss that enhances existing post-hoc methods for detecting OOD instances. We expect the observations and analyses in this work could inspire the future design of loss functions for OOD detection. Some future works include:

\emph{Theoretical understanding.} In this work, we empirically show that Logit Normalization can significantly improve OOD detection performance. On the theoretical side, we only present an analysis to show why the softmax cross-entropy loss encourages to produce logits with larger magnitudes, leading to the overconfidence issue that makes it challenging to distinguish ID and OOD examples. In the future work, we hope to provide a more rigorous theoretical justification to analyze how the LogitNorm loss improves OOD detection.

\emph{Hyperparameter tuning.} In our experiments, we tune the hyperparameter $\tau$ with a validation set -- Gaussian noises. Although the proposed method can achieve significant improvement after tuning, the tuning process is computationally expensive because it needs to train multiple models. Therefore, we expect that future work will be able to automatically adjust $\tau$ during training.

\section{Detailed Experimental Results}
\label{app:detail_results}

We report the performance of OOD detectors on each OOD test dataset in Table~\ref{tab:results_3}, \ref{tab:results_1}, and \ref{tab:results_2}. In particular, Table~\ref{tab:results_3} shows the detail performance of LogitPenalty and GODIN methods. Table~\ref{tab:results_1} shows the detail performance of different scoring functions with CE loss and LogitNorm loss. Table~\ref{tab:results_2} shows the detail performance of CE loss and LogitNorm loss with different model architectures.

\begin{table*}[h]
\centering
\caption{OOD detection performance comparison with Logit Penalty ($\lambda=0.05$) and GODIN methods. We train WRN-40-2 \cite{zagoruyko2016wide} on CIFAR-10 dataset. All values are percentages.}
\label{tab:results_3}
\renewcommand\arraystretch{1.5}
\resizebox{1.00\textwidth}{!}{
\setlength{\tabcolsep}{5mm}{
\begin{tabular}{c|ccc|ccc}
\toprule
Method & \multicolumn{3}{c}{LogitPenalty} & \multicolumn{3}{c}{GODIN} \\
\midrule
OOD dataset  & FPR95 $\downarrow$ & AUROC $\uparrow$ & AUPR $\uparrow$  & FPR95 $\downarrow$ & AUROC $\uparrow$ & AUPR $\uparrow$\\
\midrule
Texture &  62.26  &  72.00  & 87.29 &  34.90  &  92.84  &  98.41   \\
SVHN  &  62.02  &  71.87  &  87.21 &  25.17  & 95.48  &  99.04  \\
LSUN-C  &  54.71  &  66.23 & 82.29 &  15.79  &  96.88  &  99.33 \\
LSUN-R  & 51.31  &  79.24  & 90.64  & 16.82  & 97.00 & 99.40 \\
iSUN  &  51.91  &  79.70  &  91.17 &  23.25  &  95.85 & 99.15 \\
Places365  & 63.51 &70.60  & 87.09  &  35.51 & 93.39  & 98.23 \\
\bottomrule
\end{tabular}
}
}
\end{table*}

\begin{table*}[h]
\centering
\caption{OOD detection performance comparison using cross-entropy loss and LogitNorm loss. We use WRN-40-2 \cite{zagoruyko2016wide} to train on the in-distribution datasets and use softmax confidence score as the scoring function. All values are percentages. $\uparrow$ indicates larger values are better, and $\downarrow$ indicates smaller values are better. \textbf{Bold} numbers are superior results.}
\label{tab:results_1}
\renewcommand\arraystretch{1.8}
\resizebox{1.00\textwidth}{!}{
\setlength{\tabcolsep}{5mm}{
\begin{tabular}{cc|ccc|ccc}
\toprule
\multicolumn{2}{c}{ID dataset}  & \multicolumn{3}{c}{CIFAR-10} & \multicolumn{3}{c}{CIFAR-100} \\
\midrule
OOD dataset & Score  & FPR95 $\downarrow$ & AUROC $\uparrow$ & AUPR $\uparrow$  & FPR95 $\downarrow$ & AUROC $\uparrow$ & AUPR $\uparrow$\\
\midrule
 ~ & ~ & \multicolumn{6}{c}{Cross-entropy loss / \textbf{LogitNorm loss (ours)}}\\
\midrule
\multirow{3}*{Texture} & ODIN &  59.86 / \textbf{26.38}  &  76.35 / \textbf{94.66}  &  92.09 / \textbf{98.68} &   80.23 / \textbf{70.64}  &  75.60 / \textbf{78.19}  &  93.49 / \textbf{93.35}  \\
~ & Energy & 51.02 / \textbf{35.13}  &  84.60 / \textbf{92.94}  &  94.90 / \textbf{98.29} &  80.89 / \textbf{79.17}  &  75.59 / \textbf{76.94}  & 93.48 / \textbf{93.29}  \\
~ & GradNorm &  78.50 / \textbf{27.26}  &  55.15 / \textbf{93.11}  &  82.96 / \textbf{98.11}  &  86.51 / \textbf{69.07}  &  55.76 / \textbf{75.21}  &  84.28 / \textbf{91.34} \\
\midrule
\multirow{3}*{SVHN} & ODIN &  53.92 / \textbf{9.17}  &  82.98 / \textbf{98.29}  &  95.17 / \textbf{99.63} &  83.52 / \textbf{45.41}  &  79.60 / \textbf{92.59}  &  95.34 / \textbf{98.48}  \\
~ & Energy & 21.27 / \textbf{10.35}  &  95.70 / \textbf{97.73}  &  99.05 / \textbf{99.55} &  84.97/ \textbf{62.41}  &  79.41 / \textbf{89.58}  &  95.31 / \textbf{97.86}  \\
~ & GradNorm &  58.95 / \textbf{5.51}  &  76.75 / \textbf{98.91}  &  93.01 / \textbf{99.76}  &  97.7 / \textbf{39.79}  &  54.65 / \textbf{93.21}  &  86.6 / \textbf{98.57} \\
\midrule
\multirow{3}*{LSUN-C} & ODIN &  13.31 / \textbf{1.65}  &  97.14 / \textbf{99.59}  &  99.35 / \textbf{99.91} &  37.45 / \textbf{13.08}  &  93.01 / \textbf{97.66}  &  98.50 / \textbf{99.50}  \\
~ & Energy & 9.63 / \textbf{4.15}  & 98.01 / \textbf{98.64}  &  99.56 / \textbf{99.73} &  33.5 / \textbf{28.68}  & 93.74 / \textbf{95.33}  &  98.64 / \textbf{99.04}  \\
~ & GradNorm &  22.03 / \textbf{1.14}  &  93.04 / \textbf{99.70}  &  98.12 / \textbf{99.94}  &  43.43 / \textbf{8.95}  &  90.72 / \textbf{98.28}  &  97.84 / \textbf{99.63} \\
\midrule
\multirow{3}*{LSUN-R} & ODIN &  27.21 / \textbf{4.71}  & 93.52 / \textbf{98.86}  &  98.41 / \textbf{99.78} &  69.69 / \textbf{68.3}  &  83.51 / \textbf{84.78}  &  96.1 / \textbf{96.52}  \\
~ & Energy & 16.5 / \textbf{13.90}  &  96.31 / \textbf{97.31}  &  99.1 / \textbf{99.49} &  70.45 / \textbf{68.84}  &  83.57 / \textbf{85.84}  &  96.11 / \textbf{96.84}  \\
~ & GradNorm &  52.41 / \textbf{16.31}  &  77.70 / \textbf{97.18}  &  92.85 / \textbf{99.45}  &  98.34 / \textbf{83.51}  &  34.18 / \textbf{76.32}  &  76.53 / \textbf{94.15} \\
\midrule
\multirow{3}*{iSUN} & ODIN &  33.31 / \textbf{5.65}  &  92.03 / \textbf{98.79}  &  98.07 / \textbf{99.76} &  74.47 / \textbf{71.11}  &  81.01 / \textbf{83.82}  &  95.43 / \textbf{96.24}  \\
~ & Energy & 19.74 / \textbf{16.00}  &  95.69 / \textbf{97.10}  & 98.98 / \textbf{99.45} &  75.66/ \textbf{72.94}  & 80.99 / \textbf{84.48}  &  95.43 / \textbf{96.49}  \\
~ & GradNorm &  59.08 / \textbf{13.76}  &  74.64 / \textbf{97.58}  &  91.77 / \textbf{99.53}  &  99.41 / \textbf{82.51}  & 32.47 / \textbf{77.39}  &  75.58 / \textbf{94.38} \\
\midrule
\multirow{3}*{Places365} & ODIN &  54.32 / \textbf{30.12}  &  81.25 / \textbf{94.04}  &  94.33 / \textbf{98.63} &  78.93/ \textbf{80.23}  &  75.55 / \textbf{76.84}  &  93.37 / \textbf{94.04}  \\
~ & Energy & 42.75 / \textbf{35.31}  &  88.09 / \textbf{92.84}  &  96.5 / \textbf{98.30} &  79.72 / \textbf{80.73}  &  75.4 / \textbf{76.86}  &  93.34 / \textbf{94.08}  \\
~ & GradNorm &  82.86 / \textbf{42.67}  &  56.46 / \textbf{91.55}  &  85.79 / \textbf{98.04}  &  96.67 / \textbf{87.52}  &  49.29 / \textbf{68.02}  &  83.88 / \textbf{91.05} \\
\bottomrule
\end{tabular}
}
}
\end{table*}

\begin{table*}[h]
\centering
\caption{OOD detection performance comparison using cross-entropy loss and LogitNorm loss with ResNet-34 and DenseNet-BC. In-distribution dataset is CIFAR-10. All values are percentages. \textbf{Bold} numbers are superior results.}
\label{tab:results_2}
\renewcommand\arraystretch{1.8}
\resizebox{1.00\textwidth}{!}{
\setlength{\tabcolsep}{5mm}{
\begin{tabular}{c|ccc|ccc}
\toprule
Model architecture  & \multicolumn{3}{c}{ResNet-34} & \multicolumn{3}{c}{DenseNet} \\
\midrule
OOD dataset  & FPR95 $\downarrow$ & AUROC $\uparrow$ & AUPR $\uparrow$  & FPR95 $\downarrow$ & AUROC $\uparrow$ & AUPR $\uparrow$\\
\midrule
 ~ & \multicolumn{6}{c}{Cross-entropy loss / \textbf{LogitNorm loss (ours)}}\\
\midrule
Texture &  56.38 / \textbf{30.68}  &  87.82 / \textbf{94.05}  &  96.32 / \textbf{98.52} &  66.14 / \textbf{39.79}  &  85.47 / \textbf{91.46}  & 96.05 / \textbf{97.78}   \\
SVHN  &  52.34 / \textbf{4.72}  &  89.76 / \textbf{98.98}  &  97.34 / \textbf{99.79} &  57.26 / \textbf{18.66}  &  91.09 / \textbf{96.81}  &  98.16/ \textbf{99.36}   \\
LSUN-C  &  32.10 / \textbf{0.51}  &  95.64 / \textbf{99.77}  &  99.14 / \textbf{99.95}  &  32.30 / \textbf{2.73}  &  95.59 / \textbf{99.35}  & 99.12 / \textbf{99.87}  \\
LSUN-R  &  43.31 / \textbf{14.19}  &  93.41 / \textbf{97.65}  &  98.57 / \textbf{99.54}  &  43.85 / \textbf{5.41}  & 94.04 / \textbf{98.70}  &  98.82 / \textbf{99.75}  \\
iSUN  &  45.80 / \textbf{14.83}  &  92.62 / \textbf{97.47}  &  98.36 / \textbf{99.51}  & 43.08 / \textbf{5.73}  &  94.13 / \textbf{98.68}  &  98.83 / \textbf{99.74}  \\
Places365  &  56.48 / \textbf{29.98}  &  87.57 / \textbf{94.16}  &  96.50 / \textbf{98.63}  & 60.26 / \textbf{38.70}  & 88.26 / \textbf{91.98}  &  97.19/ \textbf{98.09}  \\
\bottomrule
\end{tabular}
}
}
\end{table*}

\end{document}